\documentclass[runningheads]{llncs}
\usepackage{graphicx}
\usepackage{xcolor}
\usepackage{multirow}
\begin{document}
\title{BrainNetGAN: Data augmentation of brain connectivity using generative adversarial network for dementia classification}
\titlerunning{BrainNetGAN}
% If the paper title is too long for the running head, you can set
% an abbreviated paper title here
%
\author{Chao Li$^\star$\inst{1,2} \and Yiran Wei\thanks{Equal contribution}\inst{1} \and Xi Chen\thanks{Corresponding author}\inst{3} \and  Carola-Bibiane Schönlieb\inst{4}} 
% \and
% Second Author\inst{2,3}\orcidID{1111-2222-3333-4444} \and
% Third Author\inst{3}\orcidID{2222--3333-4444-5555}}
%
\authorrunning{C. Li et al.}
% First names are abbreviated in the running head.
% If there are more than two authors, 'et al.' is used.
%
\institute{Department of Clinical Neurosciences, University of Cambridge  
\and  Shanghai General Hospital, Shanghai Jiao Tong University
\and Department of Computer Science, University of Bath \email{xc841@bath.ac.uk}
\and Department of Applied Mathematics and Theoretical Physics, \\University of Cambridge }
\maketitle              % typeset the header of the contribution
\begin{abstract}
Alzheimer's disease (AD) is the most common age-related dementia, which significantly affects an individual’s daily life and impact socioeconomics. It remains a challenge to identify the individuals at risk of dementia for precise management. Brain MRI offers a non-invasive biomarker to detect brain aging. Previous evidence shows that the structural brain network generated from the diffusion MRI promises to classify dementia accurately based on deep learning models. However, the limited availability of diffusion MRI challenges the model training of deep learning. We propose the BrainNetGAN, a variant of the generative adversarial network, to efficiently augment the structural brain networks for dementia classifying tasks. The BrainNetGAN model is trained to generate fake brain connectivity matrices, which are expected to reflect the latent distribution and topological features of the real brain network data. Numerical results show that the BrainNetGAN outperforms the benchmarking algorithms in augmenting the brain networks for AD classification tasks.

\end{abstract}
\section{Introduction}
\label{submission}

Alzheimer's disease (AD) is the most common age-related dementia that significantly impacts the cognitive performance and the socioeconomic status of patients \cite{mattson2004pathways}. The structural brain network, constructed from the diffusion MRI (dMRI), is an emerging technique to quantify the complex brain white matter structure and incorporate the prior knowledge of brain anatomy. Specifically, tractography performed on dMRI can quantify \textit{the connectivity strength} between the separated anatomical brain regions defined on the brain atlas. Numerous studies have reported the usefulness of the structural brain network in characterising a broad spectrum of neurological diseases \cite{griffa2013structural,wei2021quantifying}, including AD \cite{ajilore2014association}.

In parallel, machine learning approaches based on the structural network have shown promise in the classification tasks to distinguish AD patients from healthy controls (CN). Particularly, state-of-the-art deep learning models have been widely used in predicting dementia using end-to-end training schemes~\cite{ahmed2018neuroimaging}. A deep learning model, however, often requires a large amount of training data as well as balanced class labels for reasonable classification performance, which both are not feasible in clinical practice. 

To mitigate this challenge, recent studies employed data augmentation techniques to increase the training sample sizes. For the augmentation of traditional images, image rotation and flipping schemes are generally effective. Nevertheless, a brain network matrix cannot be simply rotated or flipped, as it could completely change the order of brain regions and introduce artifacts to the predictive model. Therefore, data augmentation approaches tailored for brain networks are desired.

\subsection{Related work}
Other data augmentation approaches have been developed to synthesize brain network matrices and adjust the imbalanced classes of the training data. Among these approaches, oversampling techniques, such as Synthetic Minority Oversampling Technique (SMOTE) \cite{chawla2002smote}, and Adaptive Synthetic Sampling (ADASYN) \cite{he2008adasyn}, are widely used. Unlike the naive augmentation methods that randomly replicate the minority samples, SMOTE generates synthetic samples by linearly interpolating two neighboring real samples. The neighborhood is identified by a standard K-nearest neighbor (KNN) approach. Developed based on the SMOTE, ADASYN produces fake samples according to the density of the class label distribution. More fake samples of minority classes are generated than the majority classes. ADASYN also adopts the KNN to cluster the samples and adjust the boundary of multiple minority classes. In recent years, ADASYN has been used to synthesize brain networks for the classification of AD ~\cite{song2019graph}. However, the KNN-assisted oversampling techniques may not effectively capture the topological property of the high dimensional brain networks, leading to significant information loss in constructing the synthetic samples. 

Generative adversarial networks (GAN)~\cite{goodfellow2014generative} is a generative model invented recently for data synthesis. It is well known in computer vision applications, where synthetic images can be indistinguishable from real images. Previous research shows that data augmentation using GAN has been successful in image classification tasks~\cite{antoniou2017data}. However, it is unclear whether the GAN can retain the topological properties of brain networks, which would require a tailored GAN architecture. 

\subsection{BrainNetGAN}
We propose a new variant of GAN with specialized architecture for conditional brain network synthesis in this work. Inspired by \cite{antoniou2017data}, we hypothesize that data augmentation could improve the performance of dementia classification tasks where brain network matrices are used as the input. The proposed GAN variant, BrainNetGAN, can generate fake brain network matrices for both AD and CN. The main contributions of this work include:

\begin{itemize}
    \item To our best knowledge, this is the first GAN variant developed specialized for synthesizing brain network matrices in dementia classification tasks.
    \item A specialized 2D convolutional kernel was applied to learn the topological property of brain networks
 \cite{kawahara2017brainnetcnn}. 
    
    % XC: we didn't specifically show any results that can clearly proof the above commented claims, I suggest to leave it to the journal version and add more justifications and supporting evidence.  
    % YW: Graph features are the reflection of topology.
    
    \item By adopting the architecture of the Wasserstein GAN with gradient penalty \cite{gulrajani2017improved}, fast and stable training of GAN was enabled on generating network matrices.
    
    \item A graph neural network (GNN) was specially adopted as the classifier to effectively evaluate the ability of algorithms in learning the topological property of brain networks.
    
\end{itemize}

Our experiments show that the proposed method outperforms the baseline and other benchmarking techniques, suggesting the advantage of using GAN for data augmentation in brain networks.

\section{Methods}

\subsection{Structural brain networks}
\label{brainnet}
\begin{figure}[ht]
\begin{center}
\centerline{\includegraphics[width=\columnwidth]{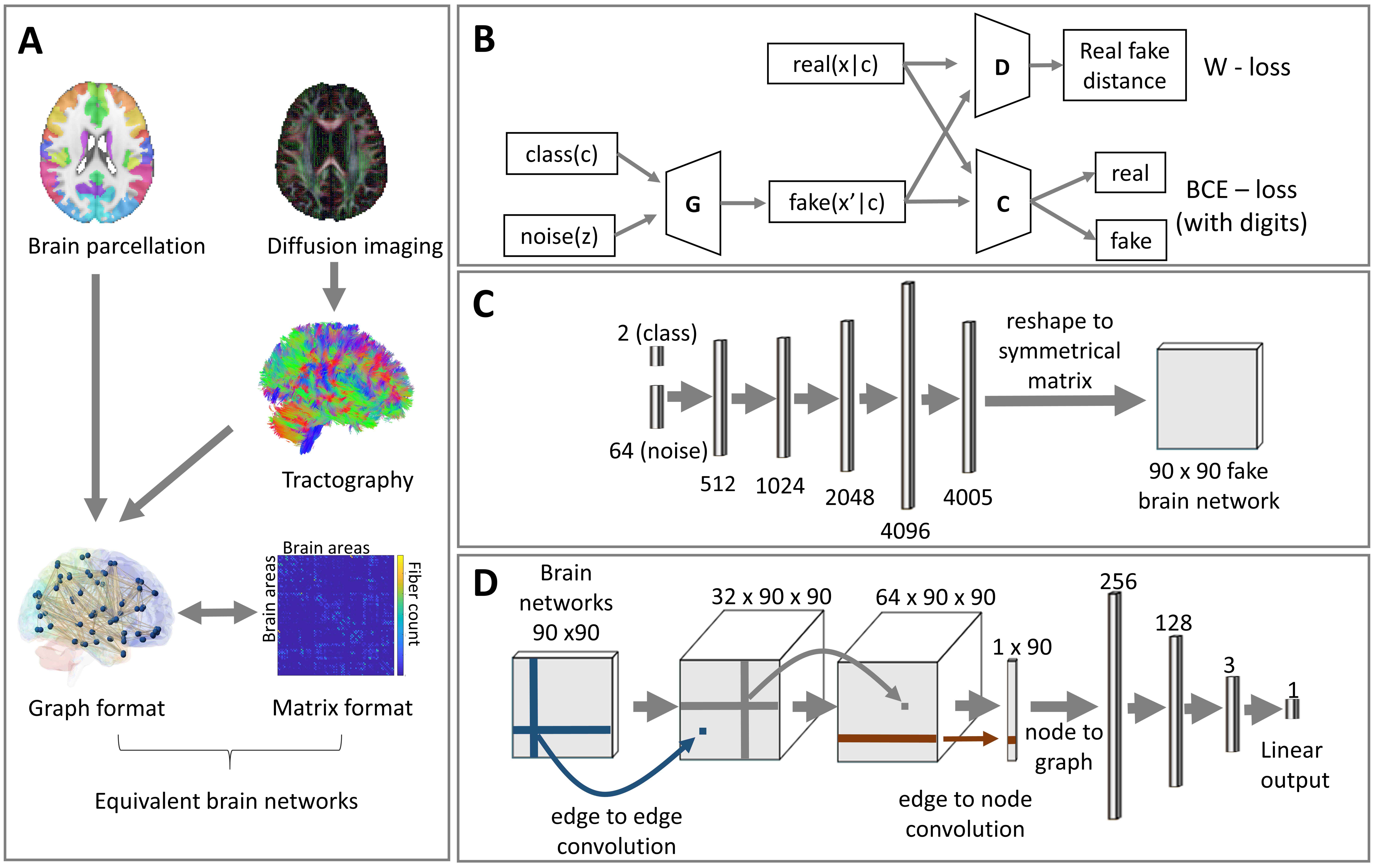}}
\caption{Architecture of the BrainNetGAN and its elements. \textbf{A}. Preparation of brain network matrices from dMRI, see Section~\ref{brainnet} for details. \textbf{B}. Design of BrainNetGAN. The framework consists of a generator $G$, a discriminator $D$ and a classifier $C$. BCE-loss: binary cross entropy loss, W-loss: Wasserstein loss. \textbf{C}. The $G$ network is a feed-forward DNN with ReLU activation functions. The numbers in \textbf{C} indicate the layer-wise input/output information. \textbf{D}. The $D$ and $C$ networks both contain special convolution kernels for the adjacency matrix of networks, see Section~\ref{dataAug} for details.}
\label{Fig:GAN}
\end{center}
\end{figure}

Adjacency matrices of the structural brain network were generated using the following steps (Figure~\ref{Fig:GAN}A). Firstly, dMRI was pre-processed in the FSL (FMRIB software library). Tractography was then performed on the processed dMRI using the Diffusion Toolkit ~\cite{mori1999three}. The grey matter regions of dMRI were divided into 90 brain regions using the Automated Anatomical Labelling (AAL) atlas, after a nonlinear registration to the standard space using Advanced Normalization Tools ~\cite{tzourio2002automated,avants2009advanced}. The structural brain networks were constructed by counting the number of tracts between each pair of brain regions to produce an adjacency matrix,  as the inputs of the following models. The tract counts of the brain network were normalized to between 0 and 1.

\subsection{Data augmentation using BrainNetGAN}
\label{dataAug}

The proposed BrainNetGAN consists of three components: a generator ($G$), a discriminator ($D$), and a classifier ($C$). The $G$ network is a feed-forward deep neural network (DNN) with four hidden layers. $G$ takes both the random vector $z$ sampled from a standard Gaussian distribution and one-hot coded class labels $c$ of the brain networks (i.e., AD or CN). The output of the $G$ network is a $1 \times 4005$ vector that is then reshaped to a $90 \times 90$ brain network matrix (with diagonal values equal to zero) as shown in Figure~\ref{Fig:GAN}C. 

Both the $D$ and $C$ networks are convolutional neural networks (CNN) consisting of three convolutional layers with specialized kernels developed by \cite{kawahara2017brainnetcnn} and five fully connected layers (Figure~\ref{Fig:GAN}D). Unlike standard convolutional kernels that only consider local neighbors of an element in a matrix, the adopted kernels take the entire row and column of the element for convolution operation, simulating both edge-edge and edge-node convolutions (e.g. a row of adjacency matrix represents all edges connecting to one node in the graph). The input of both $D$ and $C$ networks is a $90 \times 90$ matrix, either synthetic (fake) or real brain network matrices. Although the architecture of both the $D$ and the $C$ networks are identical, the loss functions are different. Specifically, a cross-entropy loss is adopted for the $C$ network to perform binary classification (AD/CN), while the Wasserstein loss is employed in the $D$ network to evaluate the difference between the real and fake network matrices.

The objective function of the BrainNetGAN consists of two components:
\begin{equation}
L_D =  E_{x \sim P_{\rm r}}  [D(x)] - E_{ \tilde{x} \sim P_{\rm g}}  [ D( \tilde{x})] + \lambda E_{\hat{x} \sim P_{\rm g}} [ (\left\| \nabla_{\hat{x}}  D(\hat{x}) \right\|_2 - 1 ) ^2]  
\label{Eq:Wgan}
\end{equation}

\begin{equation}
L_C = E_{ \tilde{x} \sim P_{\rm g}}  [\log ( C( \tilde{x}))] + E_{ x \sim P_{\rm r}}  [\log( C(x))]
\label{Eq:CE}
\end{equation}

Equation 1 represents the Wasserstein loss with gradient penalty. The first two components denote the Earth-Mover distance $W(P_r, P_g)$ between real and fake distributions that the generator tries to minimize. $P_{\rm r}$ denotes the distribution of real samples, $D$ is the discriminator, $D(x)$ represents the outcome of $D$ given input $x$, $P_{\rm g}$ denotes the distribution of fake samples defined by $\widetilde{x} = G(z|c)$, $z \sim p(z)$. $G(z|c)$ describes the generative process using a Gaussian noise $z$ conditioned on label $c$. The third component of Equation 1 represents gradient penalty that enforces the 1-Lipschitz continuity \cite{gulrajani2017improved} by penalizing the gradient norm of the random samples  $\hat{x} \sim P_{\hat{x}}$ . $\lambda$ denotes the gradient penalty coefficient. Equation 2 is the loss of classifier $C$ which represents the log-likelihood of the correct class that the generator $G$ and the classifier $C$ both try to maximize. 

\begin{figure}[ht]
\begin{center}
\centerline{\includegraphics[width=\columnwidth]{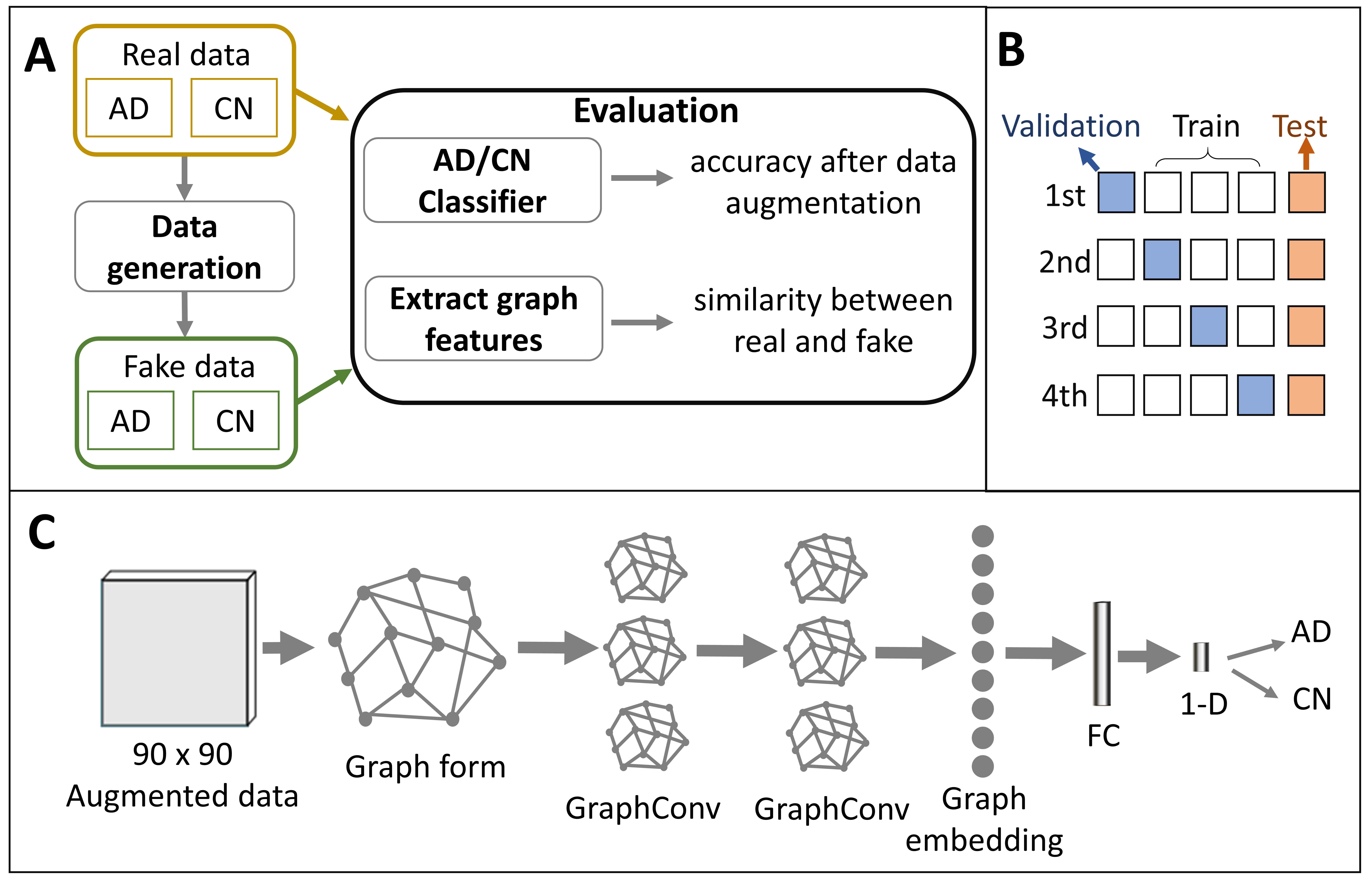}}
\caption{Components of the experiments. \textbf{A.} Overall experimental design. \textbf{A.Left}: Real data was first used to generate the same amount of fake data using different data generating methods. \textbf{A.Right}: The fake samples were used in AD/CN classification tasks to quantitatively evaluate the performance of data augmentation methods. The similarity between real and fake network matrices was also calculated using the graph features extracted from the GNN. \textbf{B.} 
An 4-fold cross-validation was used in the network training. \textbf{C.} A graph convolutional neural network was adopted as the classifier to classify AD from CN, and at the same time, evaluate (through the extracted graph features) the performance of the data augmentation methods in capturing the topological property of the brain networks.}
\label{Fig:experiment}
\end{center}
\end{figure}

\subsection{Data acquisition and experimental setup}
In this study, diffusion MRIs of 110 AD and 110 age-matched CN were obtained from the Alzheimer's Disease Neuroimaging Initiative (ADNI) (\url{adni.loni.usc.edu}).

The generated data from BrainNetGAN and baseline methods was evaluated by both classifying AD/CN and similarity metrics of graph features (Figure~\ref{Fig:experiment}A). In assessing the similarity metrics, all real data was used to generate fake data. For the classification task, real data was shuffled and split into the training, validation and testing sets with a ratio of $3:1:1$ to perform an 4-fold cross-validation (Figure~\ref{Fig:experiment}B). For each training set, three different methods produced the same sample size as the training real data.  

For BrainNetGAN, the learning rate (lr) of  $G$, $D$, and $C$ were the same during hyper-parameters tuning, the value of lr ranged between 0.00001 and 0.001. In an empirical study, 0.0005 was the large lr to guarantee the fastest convergence. The input dimension of noise $z$ was empirically tested from 64 to 256, and 64 was used in this work as it produced the most reliable convergence results.
The SMOTE and ADASYN methods were both implemented using \textit{imblearn library} \cite{JMLR:v18:16-365}.

A GNN classifier with graph convolutional layers implemented using pytorch-geometric \cite{FeyLenssen2019} was used as the AD/CN classifier to evaluate whether the generated data captured the topological differences between AD and CN networks (Figure~\ref{Fig:experiment}C). The GNN consists of graph convolutional layers called GraphConv \cite{morris2019weisfeiler}, and fully connected layers that embed the graphs. Two experiments were conducted: in the first experiment, the fake data generated by the multiple methods was fed into the classifier without the real training data. In the second experiment, the generated data was mixed with the real training data for training the classifier. The classifiers were trained for a maximum of 100 epochs with random initial weights, and the training was repeated five times to avoid statistical bias. An early stopping scheme was applied to prevent over-fitting so that the training is stopped when the validation loss stops decreasing. Average accuracy, precision, and recalls of validation and test are reported.

\begin{table}[ht]
\centering
\caption{The similarity between graph features of real and fake brain networks }
\label{tab:similarity}
%\resizebox{\columnwidth}{!}{%
\begin{tabular}{ccccc}
\hline
 & \multicolumn{2}{c}{Kullback–Leibler   Divergence} & \multicolumn{2}{c}{Maximum Mean Discrepancy} \\ \cline{2-5} 
 & CN & AD  & CN & AD \\ \hline
\multicolumn{5}{c}{Edge weight} \\ \hline
SMOTE & 0.510 & 0.482 & \textbf{0.101} & \textbf{0.119} \\
ADASYN & 0.523 & 0.533 & 0.120 & 0.132 \\
BrainNetGAN & \textbf{0.473} & \textbf{0.460} & 0.110 & 0.123 \\ \hline
\multicolumn{5}{c}{Node strength} \\ \hline
SMOTE & 0.603 & 0.613 & 0.045 & 0.054 \\
ADASYN & 0.611 & 0.587 & 0.056 & 0.059 \\
BrainNetGAN & \textbf{0.545} & \textbf{0.559} & \textbf{0.025} & \textbf{0.037} \\ \hline
\multicolumn{5}{c}{Local efficiency} \\ \hline
SMOTE & 0.428 & 0.398 & 0.035 & 0.037 \\
ADASYN & 0.523 & 0.410 & 0.042 & 0.039 \\
BrainNetGAN & \textbf{0.247} & \textbf{0.214} & \textbf{0.009} & \textbf{0.014} \\ \hline
\multicolumn{5}{c}{Global efficiency} \\ \hline
SMOTE & 0.539 & 0.484 & 0.042 & 0.064 \\
ADASYN & 0.514 & 0.543 & 0.056 & 0.064 \\
BrainNetGAN & \textbf{0.277} & \textbf{0.243} & \textbf{0.013} & \textbf{0.020} \\ \hline
\end{tabular}%
%}
\end{table}

\section{Numerical results}
\begin{table}[ht]
\centering
\caption{Evaluate data augmentation performance using graph neural networks}
\label{tab:classification}
\resizebox{\textwidth}{!}{%
\begin{tabular}{ccccccc}
\hline
 & \multicolumn{3}{c}{{Validation}} & \multicolumn{3}{c}{{Test}} \\ \cline{2-7} 
 & Accuracy & Precision & Recall & Accuracy & Precision & Recall \\ \hline
Baseline & 0.819±0.083 & 0.790±0.061 & 0.710±0.059 & 0.791±0..027 & 0.658±0.083 & 0.668±0.063 \\ \hline
\multicolumn{7}{c}{Substitute real training data for generated fake data in AD/CN classification task} \\ \hline
SMOTE & 0.819±0.069 & 0.709±0.049 & 0.790±0.068 & 0.767±0.023 & 0.725±0.096 & 0.748±0.094 \\
ADASYN & 0.793±0.075 & 0.712±0.040 & 0.705±0.087 & 0.742±0.105 & 0.702±0.115 & 0.702±0.098 \\
BrainNetGAN & \textbf{0.831±0.072} & \textbf{0.772±0.086} & \textbf{0.822±0.062} & \textbf{0.812±0.068} & \textbf{0.795±0.104} & \textbf{0.803±0.102} \\ \hline
\multicolumn{7}{c}{Combine real training data with generated fake data in AD/CN classification task} \\ \hline
Baseline ± SMOTE & 0.820±0.092 & 0.801±0.051 & 0.820±0.055 & 0.802±0.025 & 0.798±0.085 & 0.812±0.119 \\
Baseline ± ADASYN & 0.778±0.093 & 0.745±0.065 & 0.712±0.070 & 0.788±0.124 & 0.718±0.094 & 0.711±0.132 \\
Baseline ± BrainNetGAN & \textbf{0.852±0.085} & \textbf{0.805±0.091} & \textbf{0.825±0.043} & \textbf{0.829±0.058} & \textbf{0.809±0.142} & \textbf{0.825±0.113} \\ \hline
\end{tabular}%
}
\end{table}

Data augmentation performance of BrainNetGAN was evaluated using the proposed dementia classifier, and results were compared to those in the baseline dataset (no augmented data) and augmented data from different methods (Table~\ref{tab:classification}). The results show that the model with data generated by BrainNetGAN achieved higher performance in the classification compared to other models. Notably, the higher recall and precision of the model augmented by BrainNetGAN imply that the distribution of the augmented data by BrainNetGAN is less biased than other methods.

Moreover, the results on the testing set showed that the classification accuracy was improved from $0.79$ to $0.83$, when the fake samples generated by BrainNetGAN were added to the real samples, which doubled the sample size of the training dataset. 

In addition, we verify the similarity between the graph features of real and fake networks generated by different methods. Kullback–Leibler divergence and Maximum mean discrepancy were used to compare three data augmentation methods(Table~\ref{tab:similarity}). Both metrics evaluate the distance between the distribution of real and fake data, where a lower value indicates high similarity. Edge weight, node strength, local efficiency, and global efficiency were computed using the Brain Connectivity Toolbox \cite{rubinov2010complex}. We compared the similarities of those graph features calculated from the brain networks augmented by the three methods.

As shown in the Table~\ref{tab:similarity}, the data generating performance of the two methods were consistent between AD and CN. The BrainNetGAN outperformed other methods in the comparison of most topological features, indicating that the topological properties of the fake brain networks generated by BrainNetGAN approximate the real data.

\section{Discussion and conclusion}
We propose the BrainNetGAN to perform data augmentation of brain network matrices for dementia classification. Numerical results demonstrated that the BrainNetGAN outperformed the benchmarking methods and generated high-quality fake samples which effectively improved the classification performance. Future work can focus on improving the brain network edge performance in fake data generation and more deliberated analysis of different types of entries in the connectivity matrix (therefore to generate better fake samples). To conclude, GAN based brain network augmentation is a promising technique that can provide clinical values in training deep learning models for the classification of neuropsychiatric diseases.

%
% ---- Bibliography ----

\bibliography{reference.bib}
\end{document}